\documentclass{article}

\PassOptionsToPackage{numbers, compress}{natbib}

\usepackage[preprint]{neurips_2025}

\usepackage[utf8]{inputenc} 
\usepackage[T1]{fontenc}    
\usepackage{hyperref}       
\usepackage{url}            
\usepackage{booktabs}       
\usepackage{amsfonts}       
\usepackage{nicefrac}       
\usepackage{microtype}      
\usepackage{xcolor}         
\usepackage{amsmath}
\usepackage{graphicx}
\usepackage{multirow}
\usepackage{tabularx}
\usepackage{amssymb} 
\newcolumntype{R}[1]{>{\raggedleft\arraybackslash}p{#1}} 
\newcolumntype{L}[1]{>{\raggedright\arraybackslash}p{#1}} 
\newcolumntype{C}[1]{>{\centering\arraybackslash}p{#1}} 

\newcolumntype{Y}{>{\raggedright\arraybackslash}X}
\newcolumntype{Z}{>{\raggedleft\arraybackslash}X}
\newcolumntype{A}{>{\centering\arraybackslash}X}

\definecolor{myblue}{HTML}{0042B6}
\usepackage{tikz}
\usetikzlibrary{plotmarks}
\usepackage{xparse}

\NewDocumentCommand{\Plot}{ m O{} }{%
    \tikz[baseline=-0.6ex,inner sep=0pt, outer xsep=0pt] {%
        \draw[line width=1pt,#1] (0,0) -- (4mm,0);%
        \if\relax\detokenize{#2}\relax\else
            \fill[#1] (2mm,0) circle (1.5pt);%
        \fi
    }%
}

\definecolor{mygreen1}{HTML}{079199}
\definecolor{mygreen2}{HTML}{7ccba2}
\definecolor{myred}{HTML}{f0746e}
\definecolor{myyellow}{HTML}{facc8e}
\usetikzlibrary{shapes.geometric}
\NewDocumentCommand{\LegendLineX}{ m m o }{%
  \tikz[baseline=-0.5ex]{
    \draw[line width=1pt, #1] (0,0) -- (5mm,0);
    \IfNoValueF {#3} {
      \node[draw=#1, fill=#1, shape=#3, minimum size=3pt, inner sep=1pt] at (2.5mm,0) {};
    }
  }%
}

\usepackage{xspace}                    
\newcommand{\aug}{\(\times8\)\xspace aug }

\usepackage{colortbl}  

\title{LRM-1B: Towards Large Routing Model}

\author{%
 Han Li \\ 
School of Automation and Intelligent Manufacturing \\
Southern University of Science and Technology \\
Shenzhen, China \\
  \texttt{12232624@mail.sustech.edu.cn} \\
  \And
  Fei Liu \\
  Department of Computer Science \\
  City University of Hong Kong \\
  Hong Kong, China \\
  \texttt{fliu36-c@my.cityu.edu.hk} \\
  \AND
   Zhenkun Wang \\
School of Automation and Intelligent Manufacturing \\
Southern University of Science and Technology \\
Shenzhen, China \\
  \texttt{12010126@mail.sustech.edu.cn} \\
  \And
  Qingfu Zhang \\
  Department of Computer Science \\
  City University of Hong Kong \\
  Hong Kong, China \\
  \texttt{qingfu.zhang@cityu.edu.hk} \\
}

\begin{document}

\maketitle

\begin{abstract}
Vehicle routing problems (VRPs) are central to combinatorial optimization with significant practical implications. Recent advancements in neural combinatorial optimization (NCO) have demonstrated promising results by leveraging neural networks to solve VRPs, yet the exploration of model scaling within this domain remains underexplored. Inspired by the success of model scaling in large language models (LLMs), this study introduces a Large Routing Model with 1 billion parameters (LRM-1B), designed to address diverse VRP scenarios. We present a comprehensive evaluation of LRM-1B across multiple problem variants, distributions, and sizes, establishing state-of-the-art results. Our findings reveal that LRM-1B not only adapts to different VRP challenges but also showcases superior performance, outperforming existing models. Additionally, we explore the scaling behavior of neural routing models from 1M to 1B parameters. Our analysis confirms power-law between multiple model factors and performance, offering critical insights into the optimal configurations for foundation neural routing solvers.

\end{abstract}

\section{Introduction}

\begin{figure}[ht]
\centering
\includegraphics[width=0.85\linewidth]{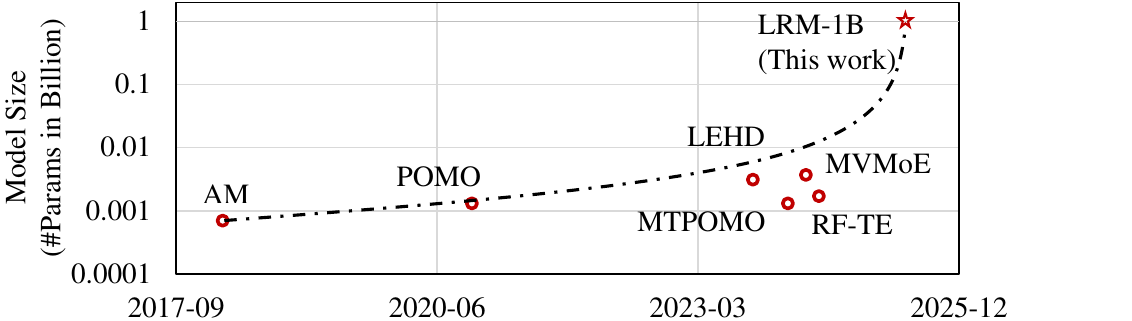}
\caption{Evolution of model size in neural routing models over time.}
\label{fig:intro}
\end{figure}

Vehicle routing problems (VRPs) are a class of combinatorial optimization problems (COPs) with significant practical importance. The objective is to plan a set of vehicle routes that meet various constraints while minimizing overall transportation cost.
In recent years, neural combinatorial optimization (NCO) has emerged as a promising approach for solving vehicle routing problems by training neural routing models \citep{BENGIO2021405,kool2018attention,bogyrbayeva2024machine}. These methods reduce reliance on handcrafted algorithmic design and benefit from modern high-performance computing hardware. Meanwhile, to meet the practical demand to simultaneously solve problems with varying node distributions, graph sizes, and multiple variants, recent research has increasingly focused on building unified models capable of addressing diverse routing tasks within a single framework \cite{zhang2022learning,xin2022generative,geisler2022generalization,wang2022a,Jiang_Wu_Cao_Zhang_2022,NEURIPS2022_ca70528f,zhou2023towards,2023arXiv230506361W,Cross_24,2024arXiv240216891L,zhou2024mvmoe,berto2024routefinder}.

Despite these advancements, the exploration of model scaling in the context of neural solvers for VRP has been relatively limited, as illustrated in Figure~\ref{fig:intro}.
For instance, the pioneering Attention Model (AM) \cite{kool2018attention} and POMO ~\cite{NEURIPS2020_f231f210} both employ a transformer-based encoder-decoder architecture, with only 0.7M and 1.3M parameters, respectively.
LEHD proposes a light encoder and heavy decoder architecture, raising the parameter count to 3.1M~\cite{luo2023neural}, and MVMoE adopts a mixture-of-experts design with a total parameter size of about 3.7M~\cite{zhou2024mvmoe}.
Recent studies have generally maintained similar model sizes, focusing instead on refining learning methods, optimizing loss functions, and developing more effective search strategies \cite{zhang2022learning,wang2022a,Jiang_Wu_Cao_Zhang_2022,NEURIPS2022_ca70528f,zhou2023towards,2023arXiv230506361W,Cross_24,2024arXiv240216891L,berto2024routefinder}. This trend overlooks the potential benefits of model scaling, which could significantly enhance the capabilities of neural solvers.

In broader AI research, particularly in the development of large language models, scaling up model size has proven to be a highly effective strategy. Studies such as those by \citet{kaplan2020scaling} and \citet{hoffmann2022training} have demonstrated power-law relationships between model size, dataset size, computational budget, and performance, suggesting that larger models tend to perform better when scaled appropriately. LLMs such as GPTs also showcase the effectiveness of large-scale models across a range of tasks \cite{brown2020language,rae2021scaling}. Inspired by these findings, our study seeks to explore the impact of model scaling within the realm of NCO for solving VRPs, aiming to uncover potential performance improvements and establish guidelines for future research in this area.

In this study, we develop a Large Routing Model with 1B parameters (\textbf{LRM-1B}), which achieves state-of-the-art (SOTA) results cross different problem variants, distribution and sizes, highlighting the potential of scaling model size. Furthermore, we train LRM with different sizes ranging from 1M to 1B parameters to investigate the scaling behavior of routing models. We reveal power-law relationships between multiple factors, such as model size, and performance, offering new insights into model and inference configuration.

The contributions of this research are outlined below:

\begin{itemize}
    \item \textbf{Development of a Large-Scale Neural Routing Model:} We have developed and trained a large-scale neural routing model, named LRM-1B, which incorporates 1 billion parameters. As illustrated in Figure~\ref{fig:intro}, this model represents a substantial scale-up in the neural solver for vehicle routing problems, aiming to leverage the increased model capacity for enhanced problem-solving capabilities.

    \item \textbf{Comprehensive Evaluation Across Diverse Scenarios:} The LRM-1B model has been evaluated across a diverse array of scenarios, encompassing different graph sizes, data distributions, and variants of vehicle routing problems. Our experiments show that LRM-1B consistently outperforms existing state-of-the-art methods for VRPs across different test scenarios.

    \item \textbf{Empirical Analysis of Scaling Laws in Neural Routing:} We have conducted an extensive empirical analysis to uncover the scaling laws associated with neural routing models. Specifically, we have examined how the performance of these models correlates with three critical factors: the model size, the number of inference trajectories, and the computational cost at inference time. The results indicate robust power-law relationships, providing valuable insights that can guide future developments in model and inference strategy optimization.
\end{itemize}



\section{Related Work}

%

\subsection{Neural Combinatorial Optimization} 
The VRP can be formulated as a sequential decision-making process, where each decision step selects the next location to visit, thereby constructing a complete route step by step. Based on this formulation, end-to-end deep learning methods have been developed to predict the next node in an autoregressive manner. Compared with supervised learning, reinforcement learning (RL) has gained significant attention, as it does not require high-quality labels for training and instead optimizes the model directly using reward signals. 

Following the success of the transformer architecture~\cite{vaswani2017attention}, several works have attempted to apply this architecture to VRP. \citet{kool2018attention} proposed the AM, a transformer-based approach trained with the REINFORCE algorithm~\cite{williams1992simple}, using a greedy rollout baseline. Building on AM, \citet{NEURIPS2020_f231f210} introduced POMO, which retains the transformer-based architecture but introduce multi-start decoding strategy. 
Specifically, for each VRP instance, the decoder performs multiple rollouts by fixing the first action at the depot and varying the second action over all feasible nodes; the best solution from these rollouts is then reported. This method significantly improves performance. 
Subsequently, numerous studies have extended transformer-based reinforcement learning models for routing problems~\cite{Xin_Song_Cao_Zhang_2021,NEURIPS2021_564127c0,Dynamic,kim2022sym}. In addition, the \aug strategy~\cite{NEURIPS2020_f231f210} is a commonly used test-time augmentation technique. It generates eight variants of a given 2D instance via coordinate transformations such as flipping and rotation. Each augmented instance is decoded independently, and the best solution among the eight is selected as the final output.
While the above approaches primarily focus on relatively simple VRP variants such as the Traveling Salesman Problem (TSP) and the Capacitated VRP (CVRP), several studies have targeted more challenging variants, including VRP with Time Windows (VRPTW)~\cite{gao2020learn,Cappart_2021,9141401,kool2022deep} and the Open VRP (OVRP)~\cite{7969477}.

\subsection{Neural Scaling Law} 
In recent years, LLMs demonstrate significant real-world value~\cite{brown2020language,team5gemini,dubey2024llama}, attracting increasing research attention. To better predict the performance of LLMs in various settings, numerous studies explore the relationships between model parameters, training dataset size, and compute budget (measured in FLOPs, for example)~\cite{kaplan2020scaling,hoffmann2022training,dubey2024llama,kumar2024scaling,allen2024physics}. 
Two pioneering studies focus on upstream cross-entropy loss \cite{kaplan2020scaling, hoffmann2022training}, empirically estimating the power-law relationships between test loss, model size, dataset size, and computational resources. 
\citet{hoffmann2022training} demonstrates that, in the absence of constraints such as dataset size or computational resources, there exists a smooth power-law relationship between model performance (typically measured by test set cross-entropy loss) and model size. Specifically, the scaling law can be approximated as $L(N) \propto N^{-\alpha_N}$, where $\alpha_N= 0.076$ is the scaling exponent. According to this equation, doubling the model size $N$ results in the loss $L$ decreasing by a factor of approximately $2^{\alpha_N}$. This relationship quantitatively links model size to performance, enabling researchers to predict the potential benefits of scaling up models based on limited empirical data.
Subsequently, many studies investigate the scaling behavior of LLMs on downstream tasks~\cite{hernandez2021scaling,gordon2021data,zhang2022examining,zhuocheng2023scaling}, the impact of post-training quantization~\cite{allen2024physics,dettmers2023case}, and extending scaling theories to the field of vision models~\cite{radford2021learning,zhai2022scaling,li2024inverse}.

\section{Large Routing Model}\label{sec:route_model}

\subsection{Datasets} 

Vehicle routing problems can be formulated on a graph $\mathcal{G}=(\mathcal{V},\mathcal{E})$, where the node set $\mathcal{V} = \{v_0, v_1,\dots, v_M\}$ consists of a depot node $v_0$ and $M$ customer nodes. The edge set $\mathcal{E} = \{(v_i,v_j) | v_i,v_j \in \mathcal{V}, i \neq j\}$ represents all possible routes between node pairs. Each node $v_i$ has a coordinate position $\vec{x}_i \in [0,1]^2$ sampled uniformly from the unit square, and an associated demand $\delta_i$. A homogeneous fleet of vehicles, each with identical load capacity $C$, is tasked with delivering goods from the depot to customers. Every customer’s demand must be satisfied exactly once, and the total demand served by any individual vehicle must not exceed its capacity. The objective is to determine the set of vehicle routes that minimizes total cost (e.g., travel distance) while respecting both demand satisfaction and other constraints.




In practical applications, a variety of specific requirements often necessitate addressing different variants of the VRPs. These variants are typically characterized by unique combinations of underlying constraints~\citep{2024arXiv240216891L,zhou2024mvmoe,berto2024routefinder}. Additionally, real-world scenarios frequently involve changes in the size of the graph and the distribution of nodes. Consequently, numerous studies have aimed to enhance the generalization capabilities of neural solvers, enabling them to effectively handle tasks that vary in size and distribution~\citep{jiang2022learning,zhou2023towards}. 



Driven by these real-world practical considerations, we train our foundation models on a mixture of problem instances varying in scale, node distribution, and VRP variants. Specifically:
\begin{itemize}
    \item \textbf{Across problem variants:} The training set incorporates 16 VRP variants (see Appendix~\ref{sec:appendix_training}). Each training batch contains a randomly selected mixture of these problem variants.
    \item \textbf{Across problem scales:} Problem scales ($M$) range from $50$ to $200$, increasing in increments of $5$ (i.e., $50, 55, 60, \dots, 200$), resulting in a total of $31$ distinct scales. For each training batch, a scale is randomly selected at the beginning of the iteration.
    \item \textbf{Across distributions:} Following \citet{zhou2023towards}, training data is generated using a mixed Gaussian distribution strategy covering $11$ different data distributions (see Appendix~\ref{sec:appendix_training} for more details). Each training batch contains a randomly selected mixture of these distributions.
\end{itemize}



\subsection{Model Structure and Configuration}
In this work, we explore the scaling behavior of routing model using the POMO framework~\cite{NEURIPS2020_f231f210}, which is a transformer-based architecture trained via RL. Four models with varying parameter sizes (1M, 5M, 40M, and 1B) are trained. Details of model configurations can be found in Table~\ref{tab:model_scale}.
\begin{table}[ht]
\caption{Configurations of models with different scales.}
\label{tab:model_scale}
\centering
\renewcommand\arraystretch{0.97}
\begin{tabularx}{\textwidth}{C{0.25\textwidth}AC{0.2\textwidth}C{0.35\textwidth}}
\toprule
Model & Layers & Attention Heads & Key/Value Embedding Dimension \\
\midrule
1.3 M         & 6      & 8            & 16            \\
5.0 M         & 12     & 16           & 16            \\
38.9 M        & 12     & 16           & 32            \\
LRM-1B 1.1 B  & 20     & 16           & 128           \\
\bottomrule
\end{tabularx}
\end{table}

\begin{figure}[ht]
\centering
\includegraphics[width=1.0\linewidth]{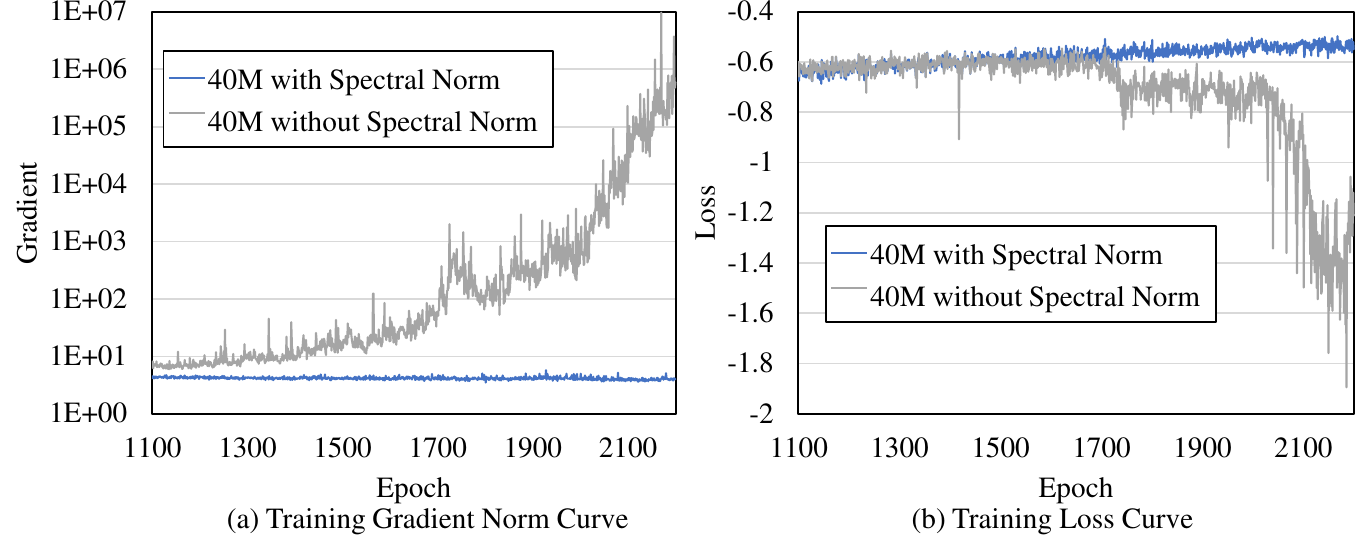}
\caption{Training curves of the 40M-parameter model. (a) Gradient norms displayed in log-scale for readability; (b) Training loss curves. Without spectral norm regularization, severe gradient fluctuations begin around $1000$ training epochs, eventually leading to gradient explosion and loss collapse. In contrast, the inclusion of spectral norm regularization effectively mitigates and controls the gradient explosion issue.}
\label{fig:spactraln}
\end{figure}

For model architecture, we follow the structure in POMO~\cite{NEURIPS2020_f231f210}, with two modifications: 1) We replace InstanceNorm with RMSNorm~\cite{zhang2019root} and adopt the SwiGLU layer~\cite{shazeer2020glu} instead of the feedforward layer, both of which are commonly used in large language models. In our setting, we find these components beneficial for model performance and convergence. 2) We apply spectral norm regularization to improve training stability. This technique constrains the spectral norm (i.e., the largest singular value) of linear layers, thereby controlling the Lipschitz constant of the network. As a result, it enhances robustness to input perturbations and stabilizes the training process. Spectral normalization has been widely adopted in the training of generative adversarial networks~\cite{yoshida2017spectral,miyato2018spectral,lin2021spectral}, RL models~\cite{bjorck2021towards}, and transformer architectures~\cite{zhai2023stabilizing}.
Figure~\ref{fig:spactraln} presents the gradient norms and training loss curves for the 40M-parameter model. Comparing runs with and without spectral norm regularization, we observe that spectral normalization effectively mitigates gradient explosion, leading to more stable and smoother convergence.
Further details of the model architecture are provided in the Appendix~\ref{sec:appendix_model}.

\subsection{Training Setup}

Every model is trained for 8,000 epochs, with each epoch consisting of 200 gradient descent steps. Regarding batch size, since memory requirements grow significantly with graph size, we adjust it dynamically for larger problems. For scales with $n > 125$, the batch size is computed as 
\begin{equation}
\text{batch size} = \left\lfloor 20 \times \left(\frac{200}{n}\right)^{2.5} \right\rfloor,
\end{equation}
where $\left\lfloor x \right\rfloor$ denotes rounding down to the nearest integer. For scales with $n \le 125$, a fixed batch size of 64 is used.

Regarding the optimizer, for models with size less than or equal to 40M, we use the Adam optimizer with a learning rate of $1 \times 10^{-4}$ and weight decay of $1 \times 10^{-6}$. For the 1B model, we switch to the Adafactor optimizer~\cite{shazeer2018adafactor} to improve memory efficiency and employ a time-dependent learning rate schedule with warm-up initialization.  For the 1B model, additional implementation details to reduce memory requirements and improve speed include mixed-precision training (FP32 and BF16), FlashAttention~\cite{dao2023flashattention}, and the Liger Kernel~\cite{hsu2024ligerkernelefficienttriton}. In addition, for all models, a shared baseline~\cite{NEURIPS2020_f231f210} and gradient clipping are applied to enhance training stability, with the latter constraining the L2 norm of the gradients to a maximum of 1.0.

All experiments are conducted on NVIDIA RTX 4090 GPUs, each with 24 GB of memory. The number of GPUs needed to train the models is: 1 for the 1 M and 5 M models, 2 for the 40 M model, and 6 for the 1 B model.


\setlength{\tabcolsep}{0pt}
\begin{table}[t]
\centering
\caption{Performance comparison across different models.}
\label{tab:compare}
\small
\renewcommand\arraystretch{0.95}
\begin{tabularx}{0.95\textwidth}{L{0.12\textwidth}|AAA|AAA|AAA}
\toprule
\multirow{2}{*}{Solver} & \multicolumn{3}{c|}{Uniform50} & \multicolumn{3}{c|}{Uniform100}  & \multicolumn{3}{c}{Uniform300} \\
\cmidrule(lr){2-4} \cmidrule(lr){5-7} \cmidrule(lr){8-10}
& Obj. & Gap & Time & Obj. & Gap & Time & Obj. & Gap & Time \\
\midrule
HGS & 11.227 & *                & 10s    & 17.964 & *                & 20s    & 35.768 & *                & 60s    \\
MTPOMO    & 11.478 & 2.318\%          & 0.002s & 18.615 & 3.829\%          & 0.005s & 38.999 & 9.862\%          & 0.117s \\
MVMoE     & 11.470 & 2.218\%          & 0.003s & 18.584 & 3.584\%          & 0.007s & 38.981 & 10.007\%         & 0.132s \\
RF-POMO   & 11.450 & 2.072\%          & 0.002s & 18.561 & 3.522\%          & 0.005s & 38.717 & 8.930\%          & 0.117s \\
RF-MoE    & 11.450 & 2.055\%          & 0.003s & 18.534 & 3.317\%          & 0.007s & 38.578 & 8.556\%          & 0.132s \\
RF-TE     & 11.447 & 2.028\%          & 0.002s & 18.485 & 3.081\%          & 0.006s & 38.722 & 9.643\%          & 0.119s \\
LRM-1B    & 11.427 & \cellcolor[HTML]{D9D9D9}{1.919\%} & 0.018s & 18.452 & \cellcolor[HTML]{D9D9D9}{2.938\%} & 0.049s & 37.403 & \cellcolor[HTML]{D9D9D9}{4.890\%} & 0.460s           \\
\midrule
\multirow{2}{*}{Solver} & \multicolumn{3}{c|}{Explosion50} & \multicolumn{3}{c|}{Explosion100}  & \multicolumn{3}{c}{Explosion300} \\
\cmidrule(lr){2-4} \cmidrule(lr){5-7} \cmidrule(lr){8-10}
& Obj. & Gap & Time & Obj. & Gap & Time & Obj. & Gap & Time \\
\midrule
HGS & 10.307 & *                & 10s    & 15.487 & *                & 20s    & 29.199 & *                & 60s    \\
MTPOMO    & 10.576 & 2.689\%          & 0.002s & 16.174 & 4.560\%          & 0.005s & 32.278 & 11.406\%         & 0.116s \\
MVMoE     & 10.568 & 2.588\%          & 0.003s & 16.136 & 4.310\%          & 0.007s & 32.243 & 11.418\%         & 0.142s \\
RF-POMO   & 10.549 & 2.439\%          & 0.002s & 16.117 & 4.201\%          & 0.005s & 32.211 & 10.749\%         & 0.121s \\
RF-MoE    & 10.551 & 2.420\%          & 0.003s & 16.096 & 4.018\%          & 0.007s & 31.954 & 9.916\%          & 0.138s \\
RF-TE     & 10.536 & 2.281\%          & 0.002s & 16.022 & 3.559\%          & 0.006s & 32.076 & 10.751\%         & 0.119s \\
LRM-1B    & 10.488 & \cellcolor[HTML]{D9D9D9}{1.945\%} & 0.017s & 15.909 & \cellcolor[HTML]{D9D9D9}{2.936\%} & 0.048s & 30.591 & \cellcolor[HTML]{D9D9D9}{5.031\%} & 0.455s \\
\midrule
\multirow{2}{*}{Solver} & \multicolumn{3}{c|}{Implosion50} & \multicolumn{3}{c|}{Implosion100}  & \multicolumn{3}{c}{Implosion300} \\
\cmidrule(lr){2-4} \cmidrule(lr){5-7} \cmidrule(lr){8-10}
& Obj. & Gap & Time & Obj. & Gap & Time & Obj. & Gap & Time \\
\midrule
HGS & 11.389 & *                & 10s    & 17.626 & *                & 20s    & 34.779 & *                & 60s    \\
MTPOMO    & 11.631 & 2.240\%          & 0.002s & 18.260 & 3.799\%          & 0.005s & 38.013 & 10.127\%         & 0.116s \\
MVMoE     & 11.624 & 2.156\%          & 0.003s & 18.238 & 3.626\%          & 0.007s & 38.021 & 10.352\%         & 0.131s \\
RF-POMO   & 11.605 & 2.009\%          & 0.002s & 18.208 & 3.496\%          & 0.005s & 37.707 & 9.092\%          & 0.117s \\
RF-MoE    & 11.607 & 1.997\%          & 0.003s & 18.190 & 3.350\%          & 0.007s & 37.600 & 8.760\%          & 0.132s \\
RF-TE     & 11.596 & 1.917\%          & 0.002s & 18.140 & 3.084\%          & 0.006s & 37.788 & 10.016\%         & 0.119s \\
LRM-1B    & 11.558 & \cellcolor[HTML]{D9D9D9}{1.671\%} & 0.017s & 18.044 & \cellcolor[HTML]{D9D9D9}{2.633\%} & 0.049s & 36.326 & \cellcolor[HTML]{D9D9D9}{4.740\%} & 0.458s \\
\midrule 
\multirow{2}{*}{Solver} & \multicolumn{3}{c|}{Rotation50} & \multicolumn{3}{c|}{Rotation100}  & \multicolumn{3}{c}{Rotation300} \\
\cmidrule(lr){2-4} \cmidrule(lr){5-7} \cmidrule(lr){8-10}
& Obj. & Gap & Time & Obj. & Gap & Time & Obj. & Gap & Time \\
\midrule
HGS & 9.823  & *                & 10s    & 14.980 & *                & 20s    & 29.344 & *                & 60s    \\
MTPOMO    & 10.101 & 2.930\%          & 0.002s & 15.695 & 4.966\%          & 0.005s & 32.414 & 11.337\%         & 0.118s \\
MVMoE     & 10.085 & 2.755\%          & 0.003s & 15.643 & 4.576\%          & 0.007s & 32.357 & 11.271\%         & 0.137s \\
RF-POMO   & 10.071 & 2.633\%          & 0.002s & 15.637 & 4.560\%          & 0.005s & 32.430 & 10.986\%         & 0.126s \\
RF-MoE    & 10.074 & 2.630\%          & 0.003s & 15.610 & 4.332\%          & 0.007s & 32.113 & 9.963\%          & 0.140s \\
RF-TE     & 10.051 & 2.402\%          & 0.002s & 15.502 & 3.629\%          & 0.006s & 32.173 & 10.675\%         & 0.119s \\
LRM-1B    & 9.997  & \cellcolor[HTML]{D9D9D9}{1.938\%} & 0.017s & 15.383 & \cellcolor[HTML]{D9D9D9}{2.915\%} & 0.049s & 30.680 & \cellcolor[HTML]{D9D9D9}{4.808\%} & 0.456s\\
\midrule
\multirow{2}{*}{Solver} & \multicolumn{3}{c|}{Grid50} & \multicolumn{3}{c|}{Grid100}  & \multicolumn{3}{c}{Grid300} \\
\cmidrule(lr){2-4} \cmidrule(lr){5-7} \cmidrule(lr){8-10}
& Obj. & Gap & Time & Obj. & Gap & Time & Obj. & Gap & Time \\
\midrule
HGS & 11.463 & *                & 10s    & 17.837 & *                & 20s    & 35.288 & *                & 60s    \\
MTPOMO    & 11.711 & 2.282\%          & 0.002s & 18.477 & 3.783\%          & 0.005s & 38.534 & 9.945\%          & 0.116s \\
MVMoE     & 11.701 & 2.154\%          & 0.003s & 18.452 & 3.583\%          & 0.007s & 38.536 & 10.130\%         & 0.134s \\
RF-POMO   & 11.684 & 2.040\%          & 0.002s & 18.423 & 3.463\%          & 0.005s & 38.227 & 8.910\%          & 0.117s \\
RF-MoE    & 11.685 & 2.008\%          & 0.003s & 18.404 & 3.305\%          & 0.007s & 38.116 & 8.592\%          & 0.132s \\
RF-TE     & 11.674 & 1.917\%          & 0.003s & 18.350 & 3.024\%          & 0.006s & 38.304 & 9.852\%          & 0.119s \\
LRM-1B    & 11.639 & \cellcolor[HTML]{D9D9D9}{1.718\%} & 0.022s & 18.266 & \cellcolor[HTML]{D9D9D9}{2.656\%} & 0.049s & 36.851 & \cellcolor[HTML]{D9D9D9}{4.682\%} & 0.458s \\
\bottomrule
\end{tabularx}
\end{table}
\setlength{\tabcolsep}{6pt}

\section{Comparison to Existing Methods}
\label{sec:main_exp}
In this section, we compare our proposed large-scale model, LRM-1B (1.1B parameters), with several state-of-the-art (SOTA) methods. Specifically, we include: 1) a traditional heuristic solver, HGS-CVRP (implemented in PyVRP), which also serves as a baseline for computing performance gaps; 2) recent learning-based approaches, including MTPOMO \cite{2024arXiv240216891L}, MVMoE \cite{zhou2024mvmoe}, and RouteFinder \cite{berto2024routefinder}. RouteFinder comprises three variants: RF-POMO, RF-MoE, and RF-TE. All these methods are capable of addressing the 16 VRP variants evaluated in this study. For MTPOMO, MVMoE, and RouteFinder, we used the publicly available pretrained models provided by the authors, including models specifically trained on graph sizes of 50 and 100. Our tests cover six scenarios with varying graph sizes from 50 to 300 nodes. For the learning-based methods (MTPOMO, MVMoE, and RouteFinder), we selected pretrained models with graph sizes closest to the tested scenarios. In contrast, LRM-1B is trained as a unified model, utilizing a single model instance for all test cases.

\subsection{Evaluation Setup}
During evaluation, we assess model performance on both in-distribution (ID) datasets with uniform node distributions and out-of-distribution (OOD) datasets with unseen distributions. The ID test sets, denoted as Uniform $M$, consist of uniformly distributed instances, where $M$ indicates the graph size. Each Uniform $M$ set comprises 16 VRP variants (Appendix~\ref{sec:appendix_training}). The OOD test sets consist of instances drawn from distributions unseen during training. Specifically, six mutation operators (Explosion, Implosion, Rotation, Linear Projection, Expansion, and Grid) introduced by \citet{bossek2019evolving} are adopted to generate unseen distributions, following the parameter configurations used in \citet{zhou2023towards}. Further details are provided in Appendix~\ref{sec:appendix_test}.
For comparison experiments, we report detailed results on four of the unseen distributions, while additional results can be found in the Appendix~\ref{sec:appendix_results}. For scaling experiments, all six unseen distributions are used, and we refer to this test set as OOD $M$.
Each OOD $M$ set comprises 16 VRP variants and 6 unseen distributions per variant. Each combination of problem scale, distribution, and variant has a test set containing 100 instances. We compute the performance gap against solutions generated by the heuristic solver HGS-CVRP~\cite{vidal2022hybrid}, implemented using the PyVRP framework~\cite{wouda2024pyvrp}. The runtime limit for instances with 50, 100, 200, and 300 nodes is set to 10s, 20s, 40s, and 60s, respectively.  

\subsection{Main Result}

Table~\ref{tab:compare} provides detailed comparative results for each method, including objective values (Obj.), performance gaps relative to the HGS (Gap), and the computation time required per instance (Time). The best-performing learning-based method in terms of solution quality is highlighted with a gray background. In addition, compared with LRM-1B (1.1B parameters), prior multi-task routing models adopt relatively small model sizes, including MTPOMO (1.29M), MVMoE (3.72M), and RF-TE (1.68M), respectively.

\begin{figure}[t]
\centering
\includegraphics[width=0.9\linewidth]{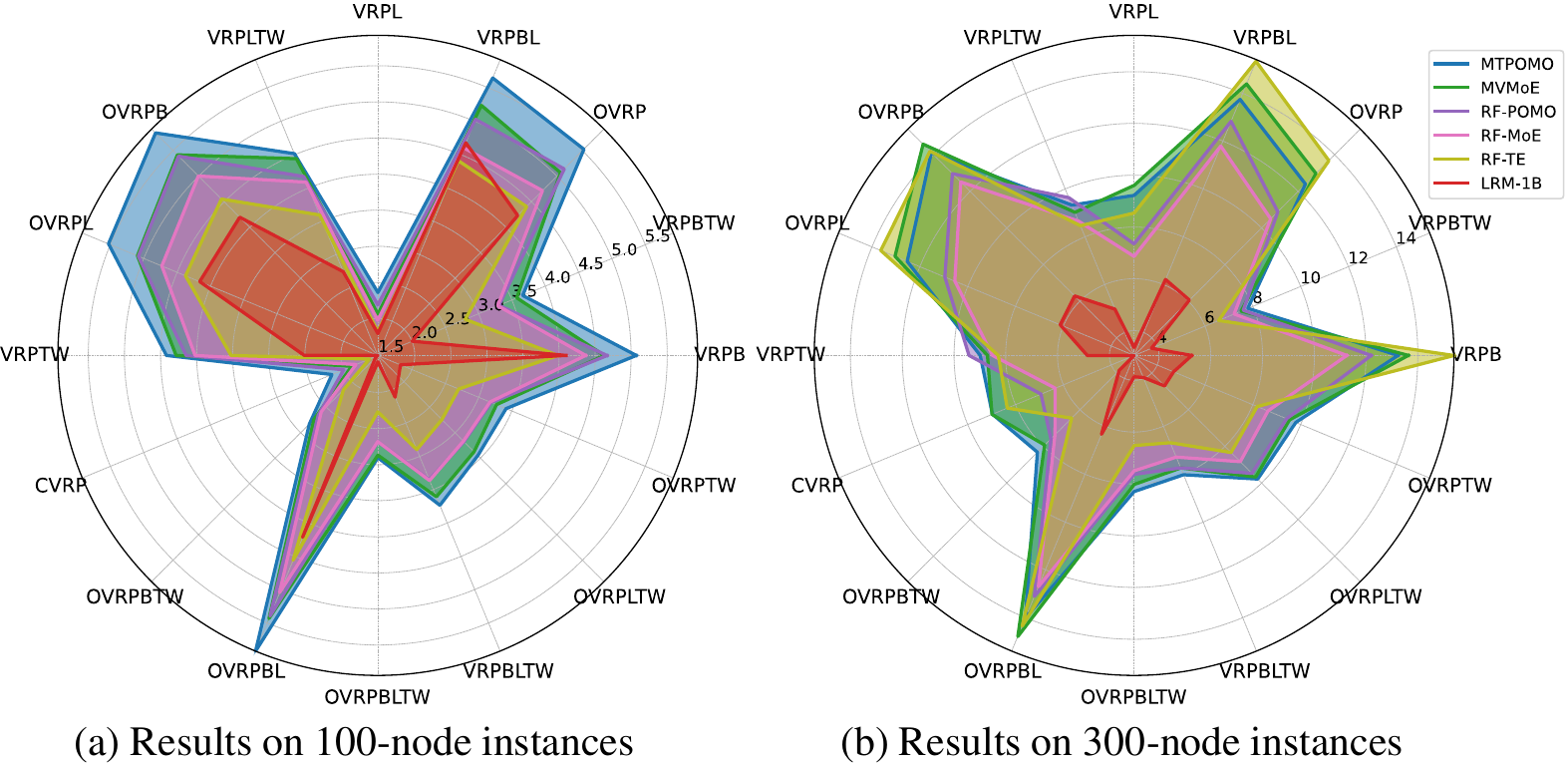}
\caption{Percentage gap (\%) of different models across VRP variants.
}
\label{fig:radar}
\end{figure}

Compared to existing multi-task learning methods, LRM-1B achieves SOTA performance across all tested scenarios, highlighting the substantial potential of scaling model sizes. Additionally, all learning-based methods outperform traditional heuristic solvers in terms of computational efficiency. Furthermore, in the Uniform50 and Uniform100 test scenarios, although other multi-task learning methods employ models specifically trained on these exact graph sizes and distributions, LRM-1B still outperforms these specialized models, despite being trained with relatively fewer samples from these particular graph sizes and distributions. This further illustrates the superior sample efficiency of LRM-1B. 
Figure~\ref{fig:radar} presents radar charts summarizing each model’s average gap, computed over five test distributions, on the various VRP variants. As shown, LRM-1B delivers consistently competitive results across all problem variants. For additional experimental results, please refer to Appendix~\ref{sec:appendix_results}.

\section{Scaling Behavior of Large Routing Model}
\label{sec:scale_exp}
In order to 1) determine whether a power-law relationship exists between model size and solver performance, and 2) examine the trade-off between inference cost and solver performance, providing practical insights for model deployment, we scale transformer-based routing solvers and conduct several empirical studies.

Specifically, our experiments consist of three main parts: 1) how the performance gap $G$ changes as model size $N$ increases; 2) how inference trajectory count $T$ and computational cost (measured in GFLOPs, $C$) affect performance gap $G$; and 3) supplementary experiments to assess the data efficiency of larger models during training. Evaluation is conducted on both ID and OOD test sets to comprehensively explore scaling behavior. Detailed results for the following scaling experiments are provided in Appendix~\ref{sec:appendix_results}.

\subsection{Scaling Law of Model Size and Performence}

\begin{figure}[t]
\centering
\includegraphics[width=1.0\linewidth]{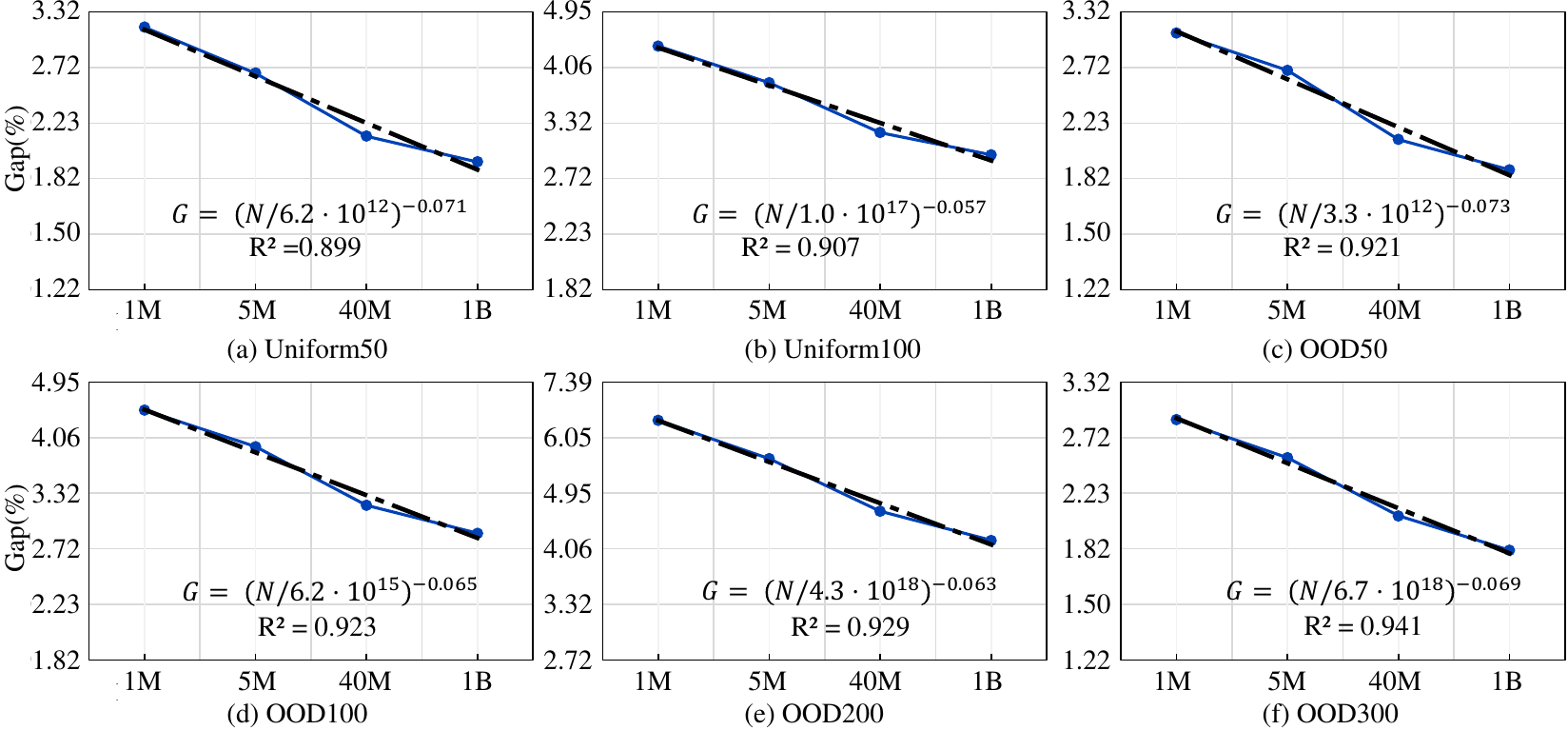}
\caption{Scaling law between model size $N$ and performance gap $G$ across various test sets. 
\Plot{myblue}[mark=*] represents actual data; 
\Plot{black, dash pattern=on 7pt off 2pt} represent power-law fits in log-log scale (i.e., $\log G$ vs. $\log N$). 
The average scaling exponent is $\alpha_N = 0.066$, implying that doubling the model size reduces the performance gap by approximately $2^{-0.066}$ ($\approx$5\% relative improvement).
}
\label{fig:scale1}
\end{figure} 

We assume a power-law relationship between model size $N$ and performance gap $G$, formulated as:
\begin{equation}
G = \left(\frac{N}{N_c}\right)^{-a_N},
\label{eq:g_n}
\end{equation}
By taking logarithms on both sides, we obtain: 
$\log(G) = -a_N \cdot \log(N) + a_N \log(N_c)$.
If the log-log plot of $G$ versus $N$ fits well to a straight line, we can conclude that $G$ and $N$ satisfy the assumed power-law relationship (Eq. \ref{eq:g_n}). Thus, Figure~\ref{fig:scale1} uses logarithmic axes to display the test performance of models with varying parameter counts. The fitted line in log-log space is obtained via ordinary least squares (OLS) regression. $R^2$ is the coefficient of determination, which ranges from 0 to 1, with values closer to 1 indicating a better fit and thus a stronger power-law relationship. In this section, evaluation is performed using the multi-start decoding and \aug strategy~\cite{NEURIPS2020_f231f210}.

Figure~\ref{fig:scale1}(a)-(f) shows a power-law relationship emerges between model size and performance gap. The average scaling exponent for all test sets is $\alpha_N = 0.066$, indicating that doubling the size of the model results in a relative improvement of approximately 5\%.

Additionally, we find that the complexity of the test scenarios influences the strength of the power-law relationship between model size $N$ and performance gap $G$. As shown in Figure~\ref{fig:scale1}(a)–(f), the task difficulty gradually increases from in-domain evaluations on seen graph sizes to more challenging out-of-domain generalization beyond the trained size. Correspondingly, $R^2$ increases, indicating that the results more closely follow the assumed power-law trend in harder scenarios. For instance, in Figure~\ref{fig:scale1}(a), the performance gain from increasing the model size from 40M to 1B shows diminishing returns. In contrast, Figure~\ref{fig:scale1}(f), which corresponds to the more difficult out-of-domain setting, shows a more consistent improvement even at the 1B scale, closely matching the fitted power-law curve.

\subsection{Scaling Laws of Inference Efficiency}

\begin{figure}[t]
\centering
\includegraphics[width=1.0\linewidth]{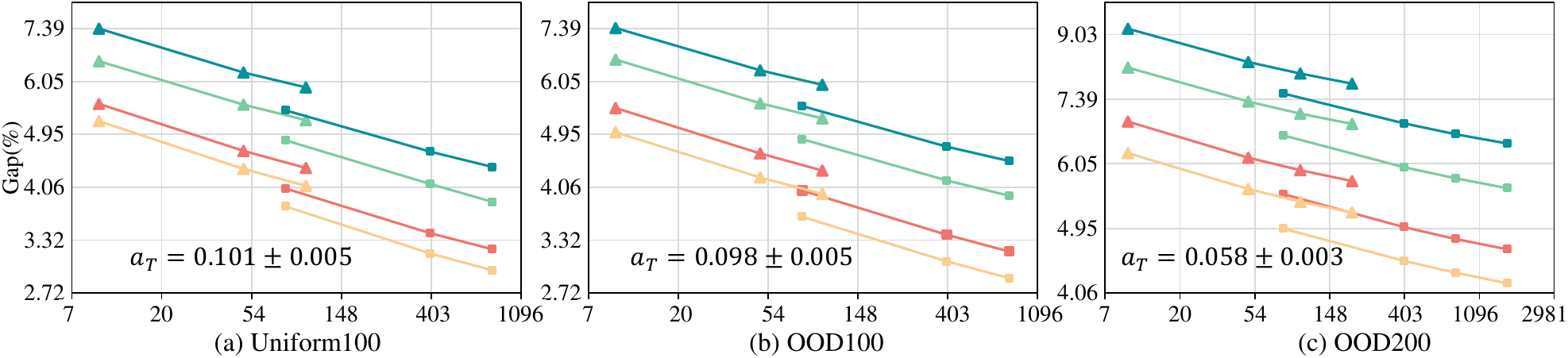}
\caption{
Scaling law between the number of inference trajectories per instance $T$ and the performance gap $G$.
Marker shapes indicate augmentation strategies: \LegendLineX{black}{}[regular polygon, regular polygon sides=3] for results without \aug, and \LegendLineX{black}{rectangle}[rectangle] for results with \aug. 
Line colors indicate model sizes:
\LegendLineX{mygreen1}{} 1M, 
\LegendLineX{mygreen2}{} 5M, 
\LegendLineX{myred}{} 40M, and \LegendLineX{myyellow}{} 1B. The exponent $a_T$ remains consistent across different model sizes and augmentation settings. Doubling the number of inference trajectories yields a relative performance improvement of $\approx$7\% on 100-node graphs and $\approx$4\% on OOD200.
}
\label{fig:scale2}
\end{figure} 

\paragraph{Scaling Law of Inference Count and Performance} 
Multi-start is a widely used decoding method, which generates $M$ trajectories by assigning each of the $M$ customer nodes as the second action. However, this approach increases inference cost by a factor of $M$. In this section, to inform practical deployment of routing solvers, we analyze how the number of inference trajectories per instance $T$ affects the performance gap $G$.

Concretely, instead of using all $M$ starting points, we select only $m$ second-action nodes, $\{1,2,\dots,m\}$, and generate $m$ trajectories. For 100-node graphs, we set $m \in \{10,50,100\}$, corresponding to the three markers on each line in Figures \ref{fig:scale2}(a)–(b), where $m=100$ corresponds to full multi-start. For 200-node graphs, we use $m \in \{10,50,100,200\}$. We further consider both without and with \aug, where \aug multiplies the trajectory count by eight.

Figure \ref{fig:scale2} plots $G$ against $T$ on log–log axes to test the scaling law 
\begin{equation}
    G \propto T^{-a_T}.
\label{eq:2}
\end{equation}
And from Figure \ref{fig:scale2}, we observe the following: 
1) $G$ and $T$ follow a power-law relationship, and the fitted exponent $a_T$ is nearly identical regardless of model size or the use of \aug, with a very small standard deviation.
2) Doubling $T$ leads to a relative improvement of approximately 7\% for 100-node graphs and 4\% for OOD200. Note that the latter is smaller than the 5\% improvement achieved by doubling the model size (as shown in Figure~\ref{fig:scale1}). This suggests that, for challenging cases such as OOD200, increasing model size is more effective than increasing the number of inference trajectories.  
3) Test‐time augmentation yields considerable gains. Under the same trajectory budget and model size, using a small number of starts combined with \aug surpasses full multi-start without augmentation.

\begin{figure}[t]
\centering
\includegraphics[width=1.0\linewidth]{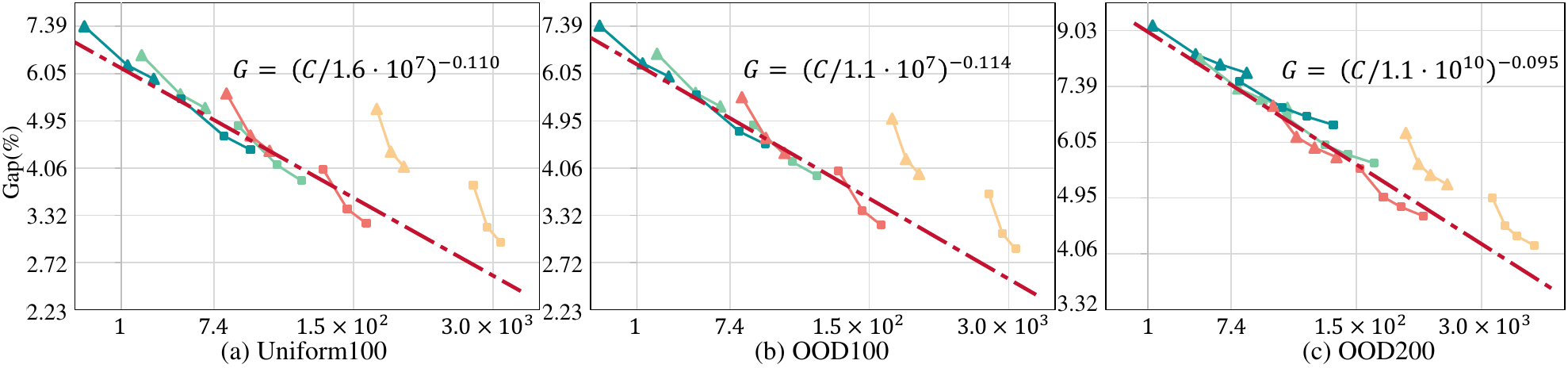}
\caption{
Scaling law between the compute cost per instance $C$ (measured in GFLOPs) and the performance gap $G$. \LegendLineX{black}{}[regular polygon, regular polygon sides=3]: results without \aug; \LegendLineX{black}{rectangle}[rectangle]: results with \aug. \LegendLineX{mygreen1}{}: 1M; \LegendLineX{mygreen2}{}: 5M; \LegendLineX{myred}{}: 40M; \LegendLineX{myyellow}{}: 1B.
The exponent $a_C$ is similar across different test sets, with an average value of $a_C = 0.106$. Doubling the inference compute cost yields a relative performance improvement of $\approx$7\% on both the 100-node and 200-node graphs.
}
\label{fig:scale3}
\end{figure} 
\paragraph{Scaling Law of Compute Cost and Performance}
We also test whether a power-law relationship exists between inference compute cost $C$ (measured in GFLOPs) and performance gap $G$: 
\begin{equation}
    G \propto C^{-a_C}.
\label{eq:3}
\end{equation}
The results are shown in Figure~\ref{fig:scale3}. Across different test cases, the fitted exponents $a_C$ are consistent, with an average value of $a_C = 0.106$. 
Additionally, we observe that the improvement in $G$ slows down when the compute cost exceeds $10^3$ GFLOPs, suggesting diminishing returns. However, when comparing OOD200 (a more challenging case) to Uniform100, we find that the marginal returns are less pronounced for Uniform100. In other words, additional compute continues to yield performance gains for harder scenarios.

\section{Conclusion}

In this paper, we introduce LRM-1B, a 1B-parameter routing model capable of solving VRP instances across various distributions, graph sizes, and problem variants. LRM-1B demonstrates consistent state-of-the-art performance compared to existing multi-task routing models across multiple benchmarks. By training models from 1M to 1B parameters, we systematically examined how performance scales with model size, the number of inference trajectories, and inference-time compute cost, uncovering power-law relationships in each case. These findings provide actionable guidance for allocating model capacity and inference budgets in practice. Overall, our work highlights the potential benefits of model scaling in routing models and offers valuable insights for optimizing trade-offs in future neural routing solver development.

\paragraph{Limitations and Future Work}
Our study focuses primarily on large models for VRPs, a specialized class of COPs. Future work could extend this framework to develop a unified model capable of solving a wider variety of combinatorial optimization tasks. Additionally, subsequent research may further explore the scaling laws of routing models during training, for example, examining the relationship between training computational cost and performance.

\begin{ack}
Use unnumbered first level headings for the acknowledgments. All acknowledgments
go at the end of the paper before the list of references. Moreover, you are required to declare
funding (financial activities supporting the submitted work) and competing interests (related financial activities outside the submitted work).
More information about this disclosure can be found at: \url{https://neurips.cc/Conferences/2025/PaperInformation/FundingDisclosure}.
Do {\bf not} include this section in the anonymized submission, only in the final paper. You can use the \texttt{ack} environment provided in the style file to automatically hide this section in the anonymized submission.
\end{ack}
\bibliographystyle{plainnat}
\bibliography{ref}
\newpage
\appendix

\section{Model Details}
\label{sec:appendix_model}
Each problem instance is represented by a node set \(\mathcal{V} = \{v_0, v_1, \dots, v_M\}\), where \(|\mathcal{V}| = M+1\). For each node,
\begin{equation}
v_i =
\begin{cases}
[\vec{x}_{i,0}, \vec{x}_{i,1}, \delta_i]^\top, & \text{if no time window constraint},\\
[\vec{x}_{i,0}, \vec{x}_{i,1}, \delta_i, t^{l}_i, t^{r}_i, t^{s}_i]^\top, & \text{else},
\end{cases}
\end{equation}
where \((\vec{x}_{i,0},\vec{x}_{i,1})\) are the coordinates, \(\delta_i\) is demand, and \(t_{\ell}, t_{r}, t_s\) denote the start and end of the time window, and $t_s$ denote the service time.

The input node set $\mathcal{V}$ is projected into a high-dimensional space to produce the initial representation $H^{(0)} \in \mathbb{R}^{(M+1) \times d_h}$. 
The encoder then refines $H^{(0)}$ through $L$ stacked layers, each comprising a multi-head self-attention (MHA) mechanism~\cite{vaswani2017attention} followed by a SwiGLU~\cite{shazeer2020glu}. The resulting representation $H^{(L)}$ serves as input for the autoregressive decoder. 

\subsection{Encoder}
Each layer includes MHA~\cite{vaswani2017attention}, SwiGLU~\cite{shazeer2020glu}, RMSNorm~\cite{zhang2019root}, and residual connections. 
\paragraph{Multi-Head Attention}
The MHA \cite{vaswani2017attention} maps queries $X$, keys $Y$, and values $Y$ into multiple subspaces and computes attention scores in parallel. Formally, it is defined as:
\begin{equation}
\text{MHA}(X, Y) = \text{Concat}(\text{head}_1, \dots, \text{head}_h) W^O,
\end{equation}
with each head computed as:
\begin{align}
\text{head}_j &= \text{Attention}(XW_j^Q, YW_j^K, YW_j^V), \\
\text{Attention}(Q,K,V) &= \text{softmax}\left(\frac{QK^\top}{\sqrt{d_k}}\right)V,
\end{align}
where $W_j^Q$, $W_j^K$, $W_j^V$, $W^O$ are learnable parameters.

\paragraph{SwiGLU}
The SwiGLU~\cite{shazeer2020glu} is a variant of gated linear units (GLU) employing the Swish function. Given an input $X$, SwiGLU is defined as:
\begin{equation}
\text{SwiGLU}(X) = (\text{Swish}(XW_1)) \odot (XW_2),
\end{equation}
where $\text{Swish}(x) = x \cdot \text{sigmoid}(x)$, and $W_1$, $W_2$ are learned linear transformations.

\paragraph{Encoder Layer}
The $i$-th layer is formulated as follows:
\begin{align}
\hat{H}^{(i)} &= \text{RMSNorm}^{(i)}\left(H^{(i-1)} + \text{MHA}^{(i)} \left(H^{(i-1)}, H^{(i-1)}\right)\right), \\
{H}^{(i)} &= \text{RMSNorm}^{(i)}\left(\hat{H}^{(i)} + \text{SwiGLU}^{(i)}(\hat{H}^{(i)})\right)
\end{align}
where ${H}^{(i-1)} \in \mathbb{R}^{(M+1) \times d_h}$ represents the node embeddings output from the $(i-1)$-th layer.

\subsection{Decoder}
After encoding, the output of the  encoder, $H^{(L)} = [\boldsymbol{h}_{0}^{(L)}, \boldsymbol{h}_{1}^{(L)}, \ldots, \boldsymbol{h}_{M}^{(L)}]$, is utilized to construct the solution. During the autoregressive decoding process, at step $t$, the context embedding is defined as:
\begin{equation}
        {H}_{c}  =\text{Concat}\big[\boldsymbol{h}^{(L)}_{{\tau}_{t}},c_{t}^{\text{l}}, c_{t}^{\text{b}}, z_{t}, l_{t}, o_{t} \big] W_{t},
\end{equation}
where ${\tau}_{t}$ is the last node of the partial solution already generated $\boldsymbol{\tau}_{t}$. The terms $c_{t}^{\text{l}}, c_{t}^{\text{b}}$ represent the remaining capacity of the vehicle for linehaul and backhaul customers, respectively. The terms $z_t$, $l_t$, and $o_t$ represent the current time, the remaining length of the current partial route (if the problem includes a length limitation), and the presence indicator of the open route, respectively. The matrix $W_c \in \mathbb{R}^{(d_h + 5) \times d_h}$ is a learnable parameter.

Then the context embeddings are processed through an MHA to generate the final query: 
\begin{equation}
    \begin{aligned}
    {q}_{c}=\text{MHA}(H_{c}^{(L)},\text{Concat}\big[\boldsymbol{h}_i^{(L)} : i \in I_t\big]),
    \end{aligned}
\end{equation}
where $I_t$ is the set of feasible actions at the current step.
The compatibility \(u_i\) is computed as:
\begin{align}
u_{i} = \begin{cases} 
\xi \cdot \tanh\left(\frac{{q}_{c} (\boldsymbol{h}_i^{(L)})^{\top}}{\sqrt{d_k}}\right) & \text{if } i \in I_t , \\
-\infty  & \text{otherwise},\\
\end{cases}
\end{align}
where $\xi$ is a predefined clipping hyperparameter. 
Finally, the action probabilities $\pi_{{\theta}}({\tau}_{g} = i \mid \mathcal{V}, \boldsymbol{\tau}_{1:g-1})$ are obtained by applying the Softmax function to $u = \{u_i\}_{i \in I_t}$.

\subsection{Feasibility Evaluation}
During each decoding step, we determine which nodes remain feasible by applying the following rules:
\begin{enumerate}
    \item No repeated visits. A customer must be served only once, and if the last visited node was the depot, the next move cannot immediately return to the depot (this prevents trivial loops).
    \item Return requirements for closed routes. In problems without an open‐route option, every tour segment must eventually return to the depot within both its time‐window and distance limits.  If visiting a candidate customer would cause the return trip (including service time) to exceed either the specified deadline or maximum travel distance, that customer is disqualified.
    \item Individual time windows. Whenever time windows apply, a customer cannot be chosen if the earliest possible arrival (plus service) would fall after its window closes.
    \item Backhaul ordering. When backhaul visits are required, they are deferred until all linehaul services are completed.  Thus, any backhaul customer is masked out as long as there remain unserved linehaul customers.
    \item Capacity checks. A node is only feasible if its demand can be loaded on the vehicle without exceeding the remaining capacity (for pickups) or the available backhaul capacity (for drop‐offs).
\end{enumerate}

The above description covers the model architecture without spectral normalization. When spectral normalization is applied, each linear layer weight matrix $W$ is replaced by $\bar W = \frac{W}{\sigma(W)}$, where $\sigma(W)$ is the largest singular value of $W$. The code is available at \url{https://anonymous.4open.science/r/LRM-1B-7B08/}.

\section{Training Setup}
\label{sec:appendix_training}

\begin{table}[t]
  \centering
  \renewcommand\arraystretch{0.95}
  \caption{16 VRP variants with five constraints.}
  \label{tab:16p}
  \begin{tabularx}{\columnwidth}{X|cccccc}
    \toprule
    & Capacity & Open Route & Backhaul & Duration Limit & Time Window \\
    \midrule
    CVRP & \checkmark & & & & \\
    OVRP & \checkmark & \checkmark & & & \\
    VRPB & \checkmark & & \checkmark & & \\
    VRPL & \checkmark & & & \checkmark & \\
    VRPTW & \checkmark & & & & \checkmark \\
    OVRPTW & \checkmark & \checkmark & & & \checkmark \\
    OVRPB & \checkmark & \checkmark & \checkmark & & \\
    OVRPL & \checkmark & \checkmark & & \checkmark & \\
    VRPBL & \checkmark & & \checkmark & \checkmark & \\
    VRPBTW & \checkmark & & \checkmark & & \checkmark \\
    VRPLTW & \checkmark & & & \checkmark & \checkmark \\
    OVRPBL & \checkmark & \checkmark & \checkmark & \checkmark & \\
    OVRPBTW & \checkmark & \checkmark & \checkmark & & \checkmark \\
    OVRPLTW & \checkmark & \checkmark & & \checkmark & \checkmark \\
    VRPBLTW & \checkmark & & \checkmark & \checkmark & \checkmark \\
    OVRPBLTW & \checkmark & \checkmark & \checkmark & \checkmark & \checkmark \\
    \bottomrule
  \end{tabularx}
\end{table}

In this section, we describe in detail how the training instances are generated.
\paragraph{Node Coordinate Distributions}  
Each instance consists of $M+1$ nodes with coordinates $\vec{x}_i \in \mathbb{R}^2$ for $i = 0, \dots, M$. Following \citet{zhou2023towards}, each training instance’s node coordinates $\{\vec{x}_i\}$ are drawn from one of 11 distributions (one uniform distribution and ten different Gaussian mixture distributions):
\begin{enumerate}
  \item Uniform. Every node coordinate is drawn from the uniform distribution $\vec{x}_i \sim U(0,1)^2$.
  \item Gaussian Mixture. This distribution is parameterized by the number of clusters $m$ and a scale factor $c$. For an instance with a Gaussian mixture distribution, the depot node is sampled as $\vec{x}_0 \sim U(0,1)^2$, and $m$ cluster centers are sampled from $U(0,c)^2$. Then the remaining $M-m$ nodes are assigned evenly to the $m$ clusters; if node $i$ belongs to cluster $j$, then $\vec{x}_i \sim \mathcal{N}(\vec{x}_j, \mathbf{I})$. Finally, all coordinates are min–max scaled to lie in $[0,1]^2$. For each instance, the pair $(m,c)$ is chosen uniformly from $\{(1,1)\}\;\cup\;\{3,5,7\}\times\{10,30,50\}$, where $(1,1)$ corresponds to a single Gaussian distribution.
\end{enumerate}


\paragraph{Capacity} For each instance, all vehicles share the same capacity $C$, and the fleet size is unlimited. Following common practice~\cite{kool2018attention,NEURIPS2020_f231f210}, we set $C = 30 + \left\lfloor \frac{M}{5} \right\rfloor$.

\paragraph{Node Demand Generation}
In the classical CVRP, every customer has a delivery (linehaul) demand. To model the backhaul variant, we allow a fraction of nodes to require pick‐up instead. Demand values are generated as follows. First, for each customer $i$, we sample a linehaul demand $\delta_i^l$ uniformly from the integers $\{1,\dots,9\}$. If the backhaul constraint is inactive, the actual demand $\delta_i$ is set to $\delta_i^l$. Otherwise, we also sample a backhaul demand $\delta_i^b$ from the same integer set. We then draw a random variable $y_i\sim U(0,1)$ and assign
\begin{equation}
\delta_i = 
\begin{cases}
\delta_i^b, & \text{if }y_i<0.2,\\
\delta_i^l, & \text{otherwise}.
\end{cases}
\end{equation}
Thus, when backhaul is enabled, each node has a 20\% chance of being a pickup customer, and an 80\% chance of remaining a delivery customer.

To improve training stability, we scale each customer’s demand by the vehicle capacity. Concretely, we compute $\delta_i' = \frac{\delta_i}{C}$, so that $\delta_i'\in[0,1]$. We then fix the (normalized) vehicle capacity at 1, ensuring that at every decoding step the remaining capacity also lies within $[0,1]$.

\paragraph{Time Windows}
For VRP variants with time window constraints, each customer $i$ (for $i=1,\dots,M$) is assigned a service time $t_i^s$ and a time window $[t_i^l, t_i^r]$. Travel speed is fixed at 1.0, and the depot’s parameters are $t_0^l = t_0^s = 0$ and $t_0^r = \mathcal{T} = 4.6$, where $\mathcal{T}$ is the total time budget for any route.

Service times $t_i^s$ are drawn uniformly from $[0.15,0.18]$, and window lengths $\Delta t_i = t_i^r - t_i^l$ are sampled from $[0.18,0.20]$. To ensure that every customer $i$ can be served on a simple trip $(0 \to i \to 0)$, we compute an upper bound for the window start:
$ e_i^{\text{up}} = \frac{\mathcal{T} - t_i^s - \Delta t_i}{d_{0i}} - 1, $
where $d_{0i}$ is the distance from the depot to customer $i$. We then sample a uniform random $y_i\in[0,1]$ and set $ t_i^l = \bigl(1 + (e_i^{\text{up}} - 1)\,y_i\bigr)\,d_{0i}$, $\quad
t_i^r = t_i^l + \Delta t_i.$
This procedure guarantees feasible time windows for all customers.

\paragraph{Distance Limit Constraint}
When a distance limit $\rho$ is imposed, every subroute must not exceed $\rho$. To ensure the simplest route $(0,i,0)$ is feasible, we draw
$
\rho \sim U\bigl(2\max_j d_{0j},\;\rho_{\max}\bigr),
$
where $d_{0j}$ is the distance from the depot to node $j$, and $\rho_{\max}=3.0$ is a fixed upper bound.

\paragraph{Summary of the 16 VRP Variants} Table~\ref{tab:16p} summarizes the 16 VRP variants used during training and evaluation.

\begin{figure}[t]
\centering
\includegraphics[width=0.9\linewidth]{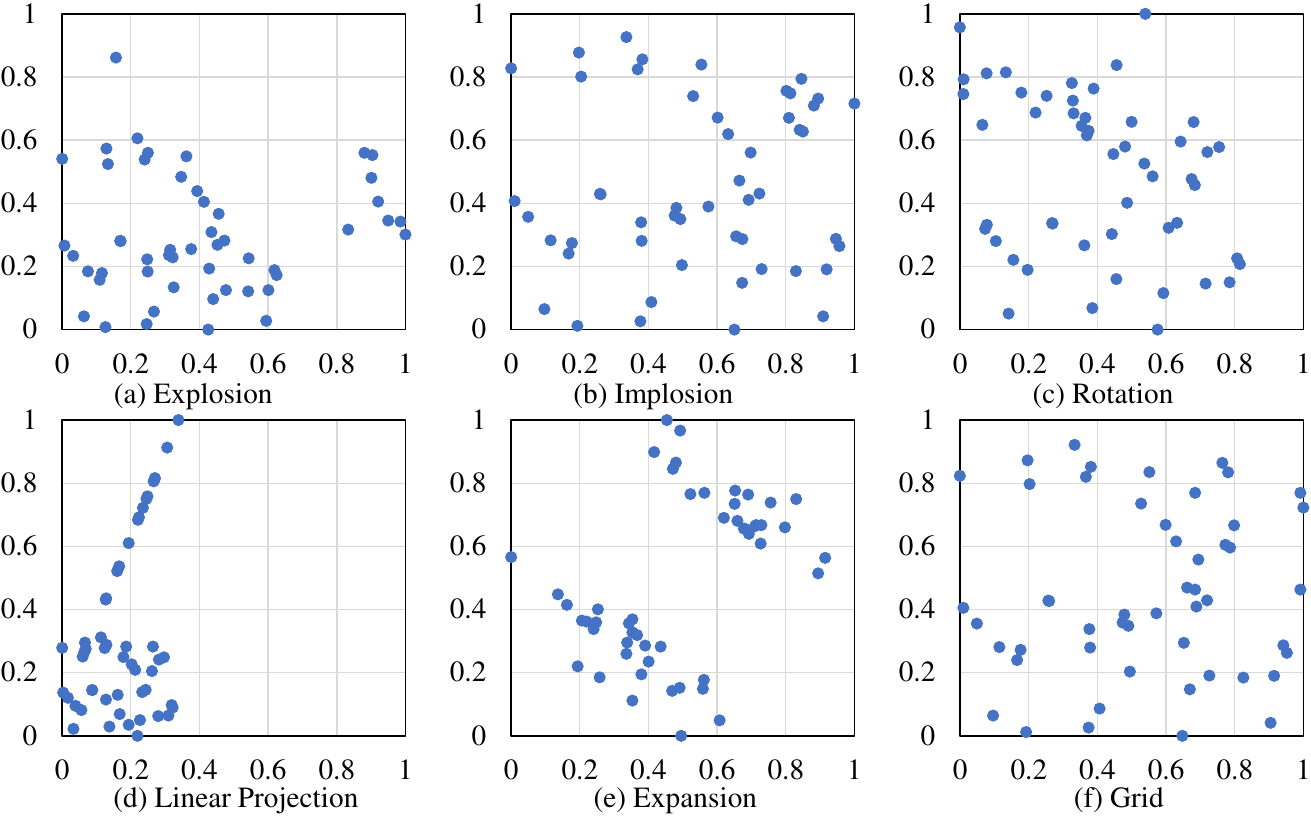}
\caption{Visualization of VRP instances with different node distributions.
}
\label{fig:dis_vis}
\end{figure} 
\setlength{\tabcolsep}{0pt}
\begin{table}[t]
\centering
\caption{Performance comparison across different models.}
\label{tab:appendix_compare}
\renewcommand\arraystretch{0.95}
\begin{tabularx}{0.95\textwidth}{L{0.12\textwidth}|AAA|AAA|AAA}
\toprule
\multirow{2}{*}{Solver} & \multicolumn{3}{c|}{Linearprojection50} & \multicolumn{3}{c|}{Linearprojection100}  & \multicolumn{3}{c}{Linearprojection300} \\
\cmidrule(lr){2-4} \cmidrule(lr){5-7} \cmidrule(lr){8-10}
& Obj. & Gap & Time & Obj. & Gap & Time & Obj. & Gap & Time \\
\midrule
HGS & 9.446 & *                & 10s    & 14.391 & *                & 20s    & 27.746 & *                & 60s    \\
MTPOMO    & 9.757 & 3.511\%          & 0.002s & 15.235 & 6.196\%          & 0.006s & 31.215 & 13.957\%         & 0.125s \\
MVMoE     & 9.751 & 3.395\%          & 0.003s & 15.206 & 5.961\%          & 0.008s & 31.191 & 13.969\%         & 0.148s \\
RF-POMO   & 9.733 & 3.247\%          & 0.002s & 15.211 & 5.908\%          & 0.006s & 31.701 & 14.311\%         & 0.133s \\
RF-MoE    & 9.740 & 3.287\%          & 0.003s & 15.181 & 5.829\%          & 0.008s & 31.028 & 12.792\%         & 0.154s \\
RF-TE     & 9.696 & 2.785\%          & 0.002s & 14.987 & 4.373\%          & 0.006s & 30.719 & 11.690\%         & 0.120s \\
LRM-1B    & 9.620 & \textbf{2.046\%} & 0.017s & 14.806 & \textbf{3.110\%} & 0.048s & 29.125 & \textbf{5.328\%} & 0.456s\\
\midrule
\multirow{2}{*}{Solver} & \multicolumn{3}{c|}{Expansion50} & \multicolumn{3}{c|}{Expansion100}  & \multicolumn{3}{c}{Expansion300} \\
\cmidrule(lr){2-4} \cmidrule(lr){5-7} \cmidrule(lr){8-10}
& Obj. & Gap & Time & Obj. & Gap & Time & Obj. & Gap & Time \\
\midrule
HGS & 9.584 & *                & 10s    & 14.174 & *                & 20s    & 26.355 & *                & 60s    \\
MTPOMO    & 9.867 & 3.031\%          & 0.002s & 14.888 & 5.121\%          & 0.005s & 29.360 & 12.226\%         & 0.118s \\
MVMoE     & 9.860 & 2.943\%          & 0.002s & 14.849 & 4.818\%          & 0.007s & 29.373 & 12.455\%         & 0.142s \\
RF-POMO   & 9.838 & 2.742\%          & 0.002s & 14.842 & 4.796\%          & 0.005s & 29.463 & 12.080\%         & 0.125s \\
RF-MoE    & 9.845 & 2.804\%          & 0.003s & 14.821 & 4.632\%          & 0.007s & 29.135 & 11.015\%         & 0.142s \\
RF-TE     & 9.821 & 2.553\%          & 0.002s & 14.738 & 4.050\%          & 0.006s & 29.225 & 11.515\%         & 0.120s \\
LRM-1B    & 9.758 & \textbf{1.961\%} & 0.017s & 14.577 & \textbf{3.006\%} & 0.048s & 27.627 & \textbf{4.997\%} & 0.454s \\
\bottomrule
\end{tabularx}
\end{table}
\setlength{\tabcolsep}{6pt}

\section{Testing Setup}
\label{sec:appendix_test}
Following \citet{bossek2019evolving}, we generate OOD datasets by applying six mutation operators to uniformly distributed instances. The six operators are defined as follows:

\begin{itemize}
    \item \textbf{Explosion:} This operator simulates a random explosion creating a cavity. A central point and radius are randomly selected, and all city nodes within this radius are displaced outward beyond the radius.
    
    \item \textbf{Implosion:} This operator represents the inverse process of the explosion. It involves randomly selecting a compression center and radius, subsequently relocating all cities within the selected radius towards the center, thereby creating dense clusters.
    
    \item \textbf{Rotation:} This operator performs rotational transformations of cities around a randomly specified pivot point. It introduces angular displacement, rearranging the spatial configuration of cities within the instance.
    
    \item \textbf{Linear Projection:} This operator projects a random subset of cities onto a randomly generated line. Specifically, it selects a subset of cities and redistributes them linearly along this generated line based on their original distances.
    
    \item \textbf{Expansion:} Combining ideas from both the explosion and linear projection operators, this operator displaces cities outward from a randomly generated line, extending the spatial layout perpendicular to the line.
    
    \item \textbf{Grid:} This operator maps randomly selected cities onto a grid-like structure. Specifically, grid width, height, and city proximity parameters are randomly determined first. A subset of cities from the instance is then repositioned onto corresponding grid points.
\end{itemize}
Detailed mathematical formulations for instance generation can be found in \citet{bossek2019evolving}. In addition, Figure~\ref{fig:dis_vis} visualizes these distributions.

\begin{table}[t]
\centering
\renewcommand\arraystretch{0.95}
\caption{Gap (\%) of different models across VRP variants with 100 nodes}
\setlength{\tabcolsep}{3pt} 
\begin{tabularx}{\textwidth}{l *{6}{>{\centering\arraybackslash}X}}
\toprule
Variant   & POMO    & MVMoE   & RF-POMO & RF-MoE  & RF-TE   & LRM-1B  \\
\midrule
VRPB     & 5.070\% & 4.625\% & 4.667\% & 4.371\% & \textbf{3.986\%} & 4.096\%          \\
VRPBTW   & 3.660\% & 3.571\% & 3.306\% & 3.288\% & 2.800\%          & \textbf{2.002\%} \\
OVRP     & 5.526\% & 5.068\% & 5.144\% & 4.718\% & 4.406\%          & \textbf{4.230\%} \\
VRPBL    & 5.647\% & 5.237\% & 5.030\% & 4.602\% & \textbf{4.409\%} & 4.676\%          \\
VRPL     & 2.359\% & 2.057\% & 2.191\% & 1.976\% & 1.818\%          & \textbf{1.791\%} \\
VRPLTW   & 4.513\% & 4.438\% & 4.173\% & 4.087\% & 3.595\%          & \textbf{2.743\%} \\
OVRPB    & 5.847\% & 5.418\% & 5.395\% & 5.004\% & 4.554\%          & \textbf{4.195\%} \\
OVRPL    & 5.531\% & 5.096\% & 5.081\% & 4.728\% & 4.373\%          & \textbf{4.160\%} \\
VRPTW    & 4.420\% & 4.285\% & 4.145\% & 4.020\% & 3.521\%          & \textbf{2.507\%} \\
CVRP     & 2.177\% & 1.881\% & 2.000\% & 1.814\% & 1.617\%          & \textbf{1.487\%} \\
OVRPBTW  & 2.832\% & 2.742\% & 2.656\% & 2.596\% & 2.162\%          & \textbf{1.536\%} \\
OVRPBL   & 5.922\% & 5.434\% & 5.395\% & 5.046\% & 4.558\%          & \textbf{4.214\%} \\
OVRPBLTW & 2.903\% & 2.869\% & 2.680\% & 2.680\% & 2.268\%          & \textbf{1.576\%} \\
VRPBLTW  & 3.734\% & 3.607\% & 3.351\% & 3.364\% & 2.898\%          & \textbf{2.112\%} \\
OVRPLTW  & 3.447\% & 3.374\% & 3.175\% & 3.167\% & 2.743\%          & \textbf{1.895\%} \\
OVRPTW   & 3.411\% & 3.268\% & 3.186\% & 3.173\% & 2.700\%          & \textbf{1.826\%} \\
\bottomrule
\end{tabularx}
\label{tab:vrp_variant_gap_100}
\end{table}
\begin{table}[t]
\centering
\renewcommand\arraystretch{0.95}
\caption{Gap (\%) of different models across VRP variants with 300 nodes}
\setlength{\tabcolsep}{3pt} 
\begin{tabularx}{\textwidth}{l *{6}{>{\centering\arraybackslash}X}}
\toprule
Variant   & POMO    & MVMoE   & RF-POMO & RF-MoE  & RF-TE   & LRM-1B  \\
\midrule
VRPB     & 13.238\% & 13.637\% & 12.199\% & 11.257\% & 15.400\% & \textbf{5.265\%} \\
VRPBTW   & 7.790\%  & 7.511\%  & 7.600\%  & 7.221\%  & 6.574\%  & \textbf{3.764\%} \\
OVRP     & 12.397\% & 12.975\% & 10.887\% & 10.502\% & 13.683\% & \textbf{6.057\%} \\
VRPBL    & 13.738\% & 14.388\% & 12.825\% & 11.791\% & 15.341\% & \textbf{6.210\%} \\
VRPL     & 9.218\%  & 9.606\%  & 7.345\%  & 6.860\%  & 8.541\%  & \textbf{3.349\%} \\
VRPLTW   & 9.321\%  & 9.019\%  & 9.635\%  & 8.653\%  & 8.466\%  & \textbf{4.971\%} \\
OVRPB    & 14.119\% & 14.578\% & 12.968\% & 12.512\% & 14.199\% & \textbf{6.283\%} \\
OVRPL    & 12.528\% & 13.047\% & 10.955\% & 10.539\% & 13.641\% & \textbf{6.125\%} \\
VRPTW    & 8.985\%  & 8.709\%  & 9.418\%  & 8.511\%  & 8.289\%  & \textbf{4.844\%} \\
CVRP     & 8.965\%  & 8.989\%  & 6.934\%  & 6.339\%  & 8.349\%  & \textbf{3.037\%} \\
OVRPBTW  & 8.324\%  & 7.933\%  & 7.588\%  & 7.363\%  & 6.457\%  & \textbf{3.851\%} \\
OVRPBL   & 14.028\% & 14.793\% & 13.113\% & 12.705\% & 14.349\% & \textbf{6.328\%} \\
OVRPBLTW & 8.296\%  & 8.023\%  & 7.631\%  & 7.495\%  & 6.514\%  & \textbf{3.859\%} \\
VRPBLTW  & 8.026\%  & 7.743\%  & 7.740\%  & 7.288\%  & 6.683\%  & \textbf{3.954\%} \\
OVRPLTW  & 9.790\%  & 9.680\%  & 9.480\%  & 8.840\%  & 8.334\%  & \textbf{4.698\%} \\
OVRPTW   & 9.800\%  & 9.541\%  & 9.417\%  & 8.639\%  & 8.183\%  & \textbf{4.686\%} \\
\bottomrule
\end{tabularx}
\label{tab:vrp_variant_gap_300}
\end{table}
\begin{table}[t]
\caption{Average performance gap (\%) across 16 VRP variants on various test sets.}
\label{tab:scaling_gap}
\centering
\renewcommand\arraystretch{1.2}
\begin{tabularx}{\textwidth}{l|XXXXXX}
\toprule
Model & Uniform50 & Uniform100 & OOD50 & OOD100 & OOD200 & OOD300 \\
\midrule
LRM-1M  & 3.141\%          & 4.379\%          & 3.076\%          & 4.478\%          & 6.439\%          & 7.884\%          \\
LRM-5M  & 2.663\%          & 3.836\%          & 2.689\%          & 3.927\%          & 5.609\%          & 6.875\%          \\
LRM-40M & 2.122\%          & 3.207\%          & 2.097\%          & 3.179\%          & 4.642\%          & 5.575\%          \\
LRM-1B  & 1.934\% & 2.960\% & 1.880\% & 2.875\% & 4.179\% & 4.927\% \\
\bottomrule
\end{tabularx}
\end{table}
\begin{table}[t]
\caption{Comparison of performance gap (\%) and compute cost (per instance) across different numbers of inference trajectories per instance (Traj.) and augmentation settings on \textbf{Uniform100}.}
\label{tab:trajectory_scaling}
\centering
\begin{tabularx}{\textwidth}{C{0.01\textwidth} C{0.08\textwidth} C{0.04\textwidth} | AA | AA | AA | AA}
\toprule
\multicolumn{3}{c|}{Uniform100} 
& \multicolumn{2}{c|}{LRM-1M}  & \multicolumn{2}{c|}{LRM-5M}  & \multicolumn{2}{c|}{LRM-40M}  & \multicolumn{2}{c}{LRM-1B} \\
$m$ & \aug & Traj. & GFLOPs & Gap & GFLOPs & Gap & GFLOPs & Gap & GFLOPs & Gap \\
\midrule
\multirow{2}{*}{10} 
&        & 10  & 0.5  & 7.388\% & 1.5  & 6.535\% & 9.6   & 5.559\% & 243.3  & 5.207\% \\
& \checkmark & 80  & 3.6  & 5.424\% & 12.4 & 4.842\% & 77.0  & 4.034\% & 1947.7 & 3.769\% \\
\midrule
\multirow{2}{*}{50} &        & 50  & 1.1  & 6.260\% & 3.6  & 5.541\% & 16.1  & 4.651\% & 329.0  & 4.344\% \\
& \checkmark & 400 & 9.2  & 4.638\% & 28.6 & 4.104\% & 129.2 & 3.408\% & 2638.1 & 3.156\% \\
\midrule
\multirow{2}{*}{100}  &        & 100 & 2.0  & 5.915\% & 6.1  & 5.225\% & 24.2  & 4.361\% & 436.3  & 4.077\% \\
& \checkmark & 800 & 16.1 & 4.379\% & 48.8 & 3.836\% & 194.5 & 3.207\% & 3501.4 & 2.960\% \\
\bottomrule
\end{tabularx}
\end{table}
\begin{table}[t]
\caption{Comparison of performance gap (\%) and compute cost (per instance) across different numbers of inference trajectories per instance (Traj.) and augmentation settings on \textbf{OOD100}.}
\label{tab:trajectory_scaling_ood100}
\centering
\begin{tabularx}{\textwidth}{C{0.01\textwidth} C{0.08\textwidth} C{0.04\textwidth} | AA | AA | AA | AA}
\toprule
\multicolumn{3}{c|}{OOD100} 
& \multicolumn{2}{c|}{LRM-1M}  & \multicolumn{2}{c|}{LRM-5M}  & \multicolumn{2}{c|}{LRM-40M}  & \multicolumn{2}{c}{LRM-1B} \\
$m$ & \aug & Traj. & GFLOPs & Gap & GFLOPs & Gap & GFLOPs & Gap & GFLOPs & Gap \\
\midrule
\multirow{2}{*}{10} 
&        & 10  & 0.4  & 7.407\% & 1.5  & 6.572\% & 9.6   & 5.471\% & 243.1  & 4.995\% \\
& \checkmark & 80  & 3.6  & 5.514\% & 12.3 & 4.864\% & 76.8  & 4.003\% & 1945.8 & 3.625\% \\
\midrule
\multirow{2}{*}{50} 
&        & 50  & 1.1  & 6.313\% & 3.5  & 5.572\% & 16.0  & 4.606\% & 327.8  & 4.209\% \\
& \checkmark & 400 & 9.1  & 4.728\% & 28.3 & 4.161\% & 128.5 & 3.385\% & 2627.0 & 3.063\% \\
\midrule
\multirow{2}{*}{100} 
&        & 100 & 2.0  & 5.973\% & 6.0  & 5.261\% & 24.0  & 4.319\% & 433.7  & 3.954\% \\
& \checkmark & 800 & 16.0 & 4.478\% & 48.3 & 3.927\% & 193.1 & 3.179\% & 3479.1 & 2.875\% \\
\bottomrule
\end{tabularx}
\end{table}

\begin{table}[t]
\caption{Comparison of performance gap (\%) and compute cost (per instance) across different numbers of inference trajectories per instance (Traj.) and augmentation settings on \textbf{OOD200}.}
\label{tab:trajectory_scaling_ood200}
\centering
\begin{tabularx}{\textwidth}{C{0.01\textwidth} C{0.08\textwidth} C{0.04\textwidth} | AA | AA | AA | AA}
\toprule
\multicolumn{3}{c|}{OOD200} 
& \multicolumn{2}{c|}{LRM-1M}  & \multicolumn{2}{c|}{LRM-5M}  & \multicolumn{2}{c|}{LRM-40M}  & \multicolumn{2}{c}{LRM-1B} \\
$m$ & \aug & Traj. & GFLOPs & Gap & GFLOPs & Gap & GFLOPs & Gap & GFLOPs & Gap \\
\midrule
\multirow{2}{*}{10} 
&        & 10   & 1.1  & 9.191\% & 3.5   & 8.149\% & 20.2  & 6.893\% & 485.2  & 6.252\% \\
& \checkmark & 80   & 9.0  & 7.519\% & 28.1  & 6.604\% & 161.6 & 5.505\% & 3883.4 & 4.949\% \\
\midrule
\multirow{2}{*}{50} 
&        & 50   & 3.1  & 8.289\% & 8.7   & 7.333\% & 35.3  & 6.165\% & 659.4  & 5.595\% \\
& \checkmark & 400  & 25.2 & 6.855\% & 70.1  & 5.986\% & 283.4 & 4.971\% & 5282.1 & 4.479\% \\
\midrule
\multirow{2}{*}{100} 
&        & 100  & 5.7  & 7.997\% & 15.3  & 7.067\% & 54.3  & 5.931\% & 877.1  & 5.377\% \\
& \checkmark & 800  & 45.4 & 6.629\% & 122.7 & 5.782\% & 435.7 & 4.793\% & 7031.7 & 4.317\% \\
\midrule
\multirow{2}{*}{200} 
&        & 200  & 10.7 & 7.752\% & 28.4  & 6.842\% & 92.2  & 5.734\% & 1312.8 & 5.201\% \\
& \checkmark & 1600 & 86.0 & 6.439\% & 227.8 & 5.609\% & 740.4 & 4.642\% & 10532.6 & 4.179\% \\
\bottomrule
\end{tabularx}
\end{table}
\setlength{\tabcolsep}{0pt}
\begin{table}[t]
\centering
\caption{Gap (\%) comparison of different models across datasets}
\begin{tabularx}{\textwidth}{C{0.03\textwidth}C{0.1\textwidth}AAAAAAA}
\toprule
Set & Size & MTPOMO & MVMoE & RF-TE & LRM-1M & LRM-5M & LRM-40M & LRM-1B \\
\midrule
A                                       & 31-79    & 3.233\%  & 3.073\%  & 2.841\%           & 2.766\%  & 2.305\%  & \textbf{2.099\%}  & 2.371\%          \\
B                                       & 30-77    & 3.797\%  & 3.888\%  & 2.581\%           & 2.573\%  & 2.638\%  & \textbf{2.050\%}  & 2.166\%          \\
F                                       & 44-134   & 11.955\% & 12.163\% & 13.009\%          & 6.205\%  & 6.237\%  & \textbf{5.485\%}  & 10.289\%         \\
M                                       & 100-199  & 5.613\%  & 5.311\%  & 5.168\%           & 5.391\%  & 4.522\%  & 4.007\%           & \textbf{3.665\%} \\
P                                       & 15-100   & 7.901\%  & 6.757\%  & 4.660\%           & 3.415\%  & 3.326\%  & \textbf{2.109\%}  & 4.637\%          \\
\multirow{4}{*}{X} & 100-300  & 7.482\%  & 6.755\%  & 5.727\%           & 6.310\%  & 5.930\%  & \textbf{5.050\%}  & 23.403\%         \\
                                        & 300-500  & 11.886\% & 11.247\% & 8.097\%           & 7.749\%  & 7.700\%  & \textbf{6.259\%}  & 19.736\%         \\
                                        & 500-700  & 24.112\% & 17.332\% & 11.281\%          & 11.669\% & 12.625\% & \textbf{10.900\%} & 21.730\%         \\
                                        & 700-1000 & 32.737\% & 19.726\% & \textbf{12.885\%} & 15.248\% & 15.761\% & 14.909\%          & 37.527\%   \\
\bottomrule
\end{tabularx} 
\label{tab:gap_comparison}
\end{table}
\setlength{\tabcolsep}{6pt}

\begin{figure}[t]
\centering
\includegraphics[width=0.9\linewidth]{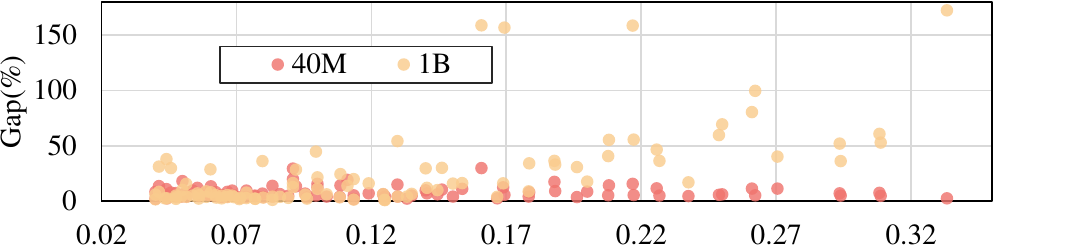}
\caption{Scatter plot of the performance gap (\%) versus average demand–capacity ratio $R$ for the 40M and 1B models on the X dataset. When the model becomes too large, the 1B variant overfits to the training‐time $R$ range (0.02–0.23), resulting in higher gaps on instances with unseen $R$ values.
}
\label{fig:appendix1}
\end{figure} 

\section{Additional Results}
\label{sec:appendix_results}
In this section, we present detailed experimental results, including those corresponding to Sections~\ref{sec:main_exp} and~\ref{sec:scale_exp}, as well as additional experiments on real-world datasets.
\paragraph{Results of Comparative Experiments}
Due to page limitations, Section~\ref{sec:main_exp} only presents results for four unseen distributions; in this section, we report the performance on the remaining two unseen distributions, Linear Projection and Expansion. The results are shown in Table \ref{tab:appendix_compare}. Overall, on these two distributions, LRM-1B achieves SOTA performance, consistent with its results on the previously presented unseen distributions. In addition, Table~\ref{tab:vrp_variant_gap_100} and Table~\ref{tab:vrp_variant_gap_300} provide detailed results corresponding to Figure~\ref{fig:radar}, reporting the performance of different routing models across VRP variants. For the 100-node instances, LRM-1B achieves the best performance on 14 out of 16 variants. For the 300-node instances, LRM-1B achieves the best performance on all 16 variants.

\paragraph{Detailed Results of Training Scaling Law} Table~\ref{tab:scaling_gap} shows the detailed results for Figure~\ref{fig:scale1}. The power-law relationship between model size and performance is derived from these data. Overall, model performance improves as model size increases across all test sets, and there is a power-law relationship between model size and performance.

\paragraph{Detailed Results of Inference Scaling Law} Tables~\ref{tab:trajectory_scaling}, \ref{tab:trajectory_scaling_ood100}, and \ref{tab:trajectory_scaling_ood200} present detailed results corresponding to Figures~\ref{fig:scale2} and \ref{fig:scale3}. The power-law relationship between the number of inference trajectories, test-time computational cost, and performance is derived from these data.

\paragraph{Data Efficiency of Large Models}
\begin{figure}[t]
\centering
\includegraphics[width=0.7\linewidth]{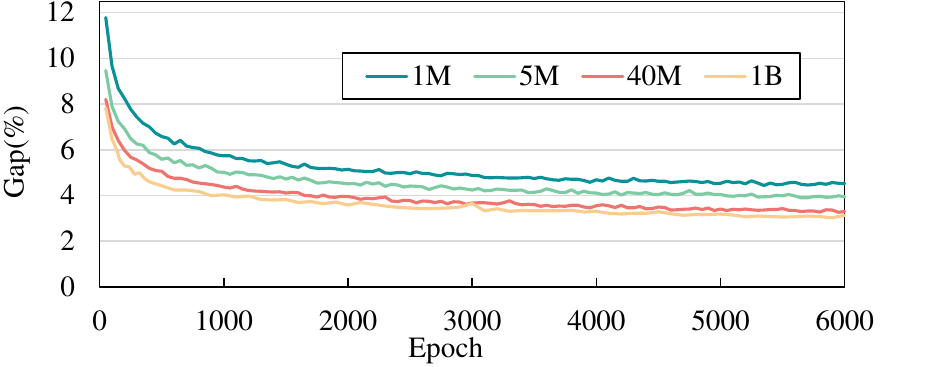}
\caption{Test performance gap (\%) on the Uniform100 test set during training. Larger models converge faster and achieve better final performance under the same number of training epochs, indicating improved data efficiency.
}
\label{fig:scale4}
\end{figure} 
Figure~\ref{fig:scale4} demonstrates the performance of models of various sizes on the Uniform100 test set during training. The results show that, with the same number of gradient descent steps, larger models converge faster, demonstrating higher data efficiency during training.

\paragraph{Real-world Dataset} In this section, we evaluate the performance of various routing models on real-world benchmarks. We consider six test suites from the CVRPLib benchmark\footnote{\url{http://vrp.atd-lab.inf.puc-rio.br/}}. For the baseline methods: MTPOMO, MVMoE, and RF-TE, we use the versions trained on 100-node instances. The results are summarized in Table~\ref{tab:gap_comparison}. Our 40M parameter model achieves SOTA results on nearly all test suites, demonstrating the benefits of moderate model scaling. In contrast, the 1B parameter model performs worse than the 40M model.

To understand this drop in performance, we compute
\begin{equation}
    R = \frac{1}{M}\sum_{i=1}^{M}\frac{\delta_i}{C},
\end{equation}
where $C$ is the vehicle capacity and $\delta_i$ is the demand of customer $i$. A larger $R$ implies that each vehicle can serve fewer customers before reaching capacity,  shortening legal subtours. During training, $R$ ranged from 0.02 to 0.23. Figure~\ref{fig:appendix1} plots $R$ versus gap for both our 40 M and 1 B models on the X dataset. We observe that the 40 M model generalizes well across all $R$ values, whereas the 1 B model’s gap increases sharply as $R$ grows, suggesting that overly large models may overfit to the training-time demand–capacity ratio.

\section{Licenses}
\label{sec:licenses}

\begin{table}[t]
\centering
\caption{List of licenses for the codes and datasets we used in this work}
\label{appendix: Licenses}
\resizebox{0.99\textwidth}{!}{%
\begin{tabular}{l |l | l | l }
\toprule
 Resource   &   Type  &  Link  & License    \\
\midrule
HGS \citep{vidal2022hybrid} & Code & https://github.com/chkwon/PyHygese & MIT License\\ 
POMO \citep{NEURIPS2020_f231f210}& Code & https://github.com/yd-kwon/POMO & MIT License \\
MVMoE \citep{zhou2024mvmoe} & Code &https://github.com/RoyalSkye/Routing-MVMoE & Available online \\
RF-TE \citep{berto2024routefinder} & Code &https://github.com/ai4co/routefinder & Available online \\
\bottomrule
\end{tabular}%
}
\end{table}

The licenses for the codes and the datasets used in this work are listed in Table \ref{appendix: Licenses}.

\end{document}